\RequirePackage{fix-cm}
\documentclass[review]{elsarticle}          
\journal{Robotics and Autonomous Systems}
\usepackage{graphicx}
\usepackage[T1]{fontenc}
\usepackage[utf8]{inputenc}

\usepackage{hyperref}       
\usepackage{url}            
\usepackage{booktabs}       
\usepackage{amsfonts}       
\usepackage{nicefrac}       
\usepackage{microtype}      

\usepackage{epstopdf}
\usepackage{subfigure}
\usepackage{enumitem}

\usepackage{xcolor}

\usepackage{amsmath}
\usepackage{multirow}
\begin{document}
\begin{frontmatter}
\title{
3D Human Pose Estimation Based on 2D-3D Consistency with Synchronized Adversarial Training
}

\author[a]{Yicheng Deng}
\ead{19120344@bjtu.edu.cn}
\author[b]{Cheng Sun}
\ead{sun.cheng.736@s.kyushu-u.ac.jp}
\author[a]{Yongqi Sun\corref{cor1}}
\cortext[cor1]{Corresponding author}
\ead{yqsun@bjtu.edu.cn}
\author[a]{Jiahui Zhu}
\ead{20120461@bjtu.edu.cn}
\affiliation[a]{organization={Key Lab of Big Data \& Artificial Intelligence in Transportation, Ministry of Education, School of Computer and Information Technology, Beijing Jiaotong University},
postcode={100044},
city={Beijing},
country={China}}
\affiliation[b]{organization={School of Information Science and Electrical Engineering, Kyushu University},
postcode={8190395},
city={Fukuoka},
country={Japan}}

\date{Received: date / Accepted: date}

\begin{abstract}
3D human pose estimation from a single image is still a challenging problem despite the large amount of work that has been performed in this field. Generally, most methods directly use neural networks and ignore certain constraints (e.g., reprojection constraints, joint angle, and bone length constraints). While a few methods consider these constraints but train the network separately, they cannot effectively solve the depth ambiguity problem. In this paper, we propose a GAN-based model for 3D human pose estimation, in which a reprojection network is employed to learn the mapping of the distribution from 3D poses to 2D poses, and a discriminator is employed for 2D-3D consistency discrimination. We adopt a novel strategy to synchronously train the generator, the reprojection network and the discriminator. Furthermore, inspired by the typical kinematic chain space (KCS) matrix, we introduce a weighted KCS matrix and take it as one of the discriminator’s inputs to impose joint angle and bone length constraints. The experimental results on Human3.6M show that our method significantly outperforms state-of-the-art methods in most cases.
\end{abstract}


\end{frontmatter}

\section{Introduction}
\label{intro}
3D human pose estimation from monocular images has always been a problem in computer vision. It can be applied in multiple fields, such as motion recognition, virtual reality, and human-computer interaction. Over the past three decades, there has been a dramatic increase in the field of 3D human pose estimation. Various methods can be divided into two main categories. One category is the end-to-end method, which directly processes the image and estimates the 3D pose through a deep learning method \cite{2016Marker, 2018OriNet, 2017VNect, 20173D, 2017self-supervised, yang_3d_2018}.
The other is the two-stage method, which first obtains the 2D joint coordinates from an image\cite{2016Human, 2017Realtime, carreira_human_2016, 2016Stacked, 2016DeepCut, 2019Deep, toshev_deeppose_2014} and then estimates the 3D pose according to the 2D joint coordinates \cite{KIM2020107462, 2018Unsupervised, 2017A, 2019RepNet}.

Although many works focus on the second stage of the two-stage method and have achieved good performance, they often ignore the fact that a well-estimated 3D pose should be able to be reprojected back to a plausible 2D pose. Recently, some methods \cite{LI2020BMVC,2019RepNet} have considered the reprojection loss and achieved better results. 
However, in their method, they trained the reprojection network by comparing the 2D reprojection with the input 2D poses, as shown in Figure \ref{diff}(a). This training strategy leads to a significant impact on the accuracy of 2D reprojection when utilizing detected 2D poses as input instead of ground truth poses.
These methods do not consider the fact that a reasonable 3D pose should not collapse when viewed at a random angle, that is, a high-quality 3D human pose estimation should look like a natural human body from any viewing angle. Hence they fail to effectively tackle the biggest challenge in 3D human pose estimation from a single image: the depth ambiguity problem. 

\begin{figure*}[tbp]
\centering
\includegraphics[width=0.8\textwidth]{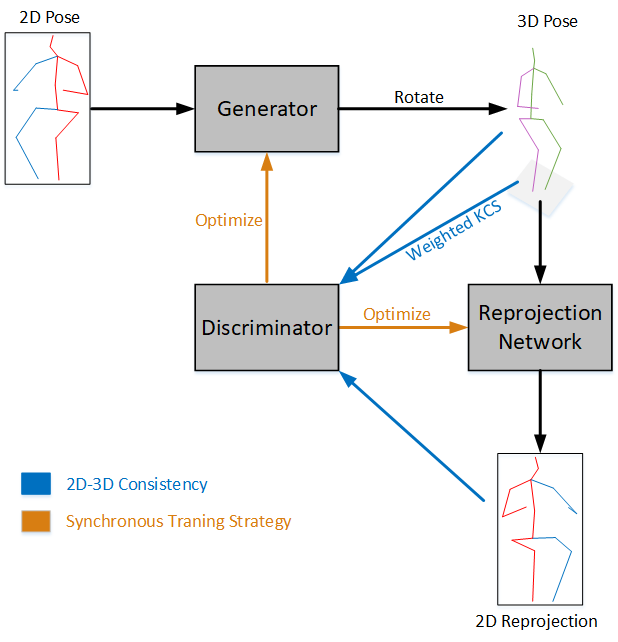}
\caption{Overview of our model. We feed the estimated 3D pose and 2D reprojection to the discriminator simultaneously, and train the generator, reprojection network, and discriminator synchronously.}
\label{abstract}
\end{figure*}

To solve the above problems, in this paper, we propose a weakly-supervised adversarial training method with 2D-3D consistency to estimate a 3D human pose, which adopts a synchronous training strategy to train the reprojection network, as shown in Figure \ref{abstract}. 
In our model, we synchronously train the generator, the reprojection network, and the discriminator using the loss function of the generative adversarial network, to make the reprojection network learn the mapping of the distribution from 3D poses to 2D poses. This training strategy can effectively reduce the influence of errors in the input of the generator. 
In our method, we first set up a generator to generate the depth (the $z$-component of the 3D coordinates) with the input 2D joint coordinates. Then, we rotate the generated 3D pose, and add a reprojection network to reproject the rotated 3D pose. 
Finally, we construct a discriminator, whose input contains a generated 3D pose and a 2D reprojection, to perform 2D-3D consistency discrimination. By combining the 2D-3D consistency with the random rotation of the generated 3D pose, our discrimination method can effectively reduce the depth ambiguity. Moreover, inspired by the original KCS matrix \cite{2019RepNet}, we distinguish the importance of the joint angle information between bones at different distances, and propose to transform the 3D pose to a more elaborate weighted kinematic chain space and make it the discriminator's third part of the input to impose constraints on bone lengths and joint angles. By using the synchronous training strategy, considering the random rotation discrimination, 2D-3D consistency discrimination and the limitations on bone lengths and joint angles, our model obtains more accurate and interpretable estimation results.
Our contributions are as follows:

\begin{itemize}
	\item We propose a novel adversarial training network specifically for 2D-3D consistency discrimination, to improve the accuracy of 3D human pose estimation.
    \item We employ a synchronous strategy to train the networks to learn the mapping of the distribution from 2D poses to 3D poses and inverse mapping, which can effectively reduce the influence of errors in the input 2D poses.
	\item We propose an elaborate improved kinematic chain space that transforms a 3D pose into a weighted kinematic chain space (wKCS) to impose constraints on bone lengths and joint angles.
\end{itemize}

The experimental results on three public datasets, Human3.6M \cite{2014Human3}, MPI-INF-3DHP \cite{2017Monocular}, and MPII \cite{2014Human}, show that our method significantly outperforms state-of-the-art weakly supervised methods and can be generalized to unknown 3D human poses.

\section{Related work}
\label{relatedwork}
At present, due to large-scale datasets for supervised training and powerful deep neural networks, significant progress has been made in 3D human pose estimation from a monocular image. We can summarize the estimation methods into two categories: end-to-end methods and two-stage methods. The end-to-end methods estimate the 3D joint point position directly from a monocular image and compare it with the 3D annotation to optimize the network \cite{2016Marker, 2018OriNet, 2017VNect, 20173D, 2017self-supervised, yang_3d_2018}. The two-stage methods divide the pose estimation into two stages. In the first stage, 2D pose detection is performed on a single image, and its 2D joint coordinates are predicted \cite{2016Human, 2017Realtime, carreira_human_2016, 2016Stacked, 2016DeepCut, 2019Deep, toshev_deeppose_2014}; in the second stage, 3D joint coordinates from the 2D joint coordinates are predicted through regression analysis or model fitting \cite{KIM2020107462, 2018Unsupervised, 2017A, 2019RepNet}. In this paper, we focus on the second stage of the two-stage methods.

From another perspective, we can further classify 3D estimation methods into three classes: fully supervised methods, unsupervised methods, and weakly supervised methods. In recent years, the GAN proposed by Goodfellow et al.\cite{goodfellow_i_generative_2014} has been a major hit in deep learning, and its application in 3D human pose estimation is also quite extensive.

\subsection{Fully supervised methods}
There have been several fully supervised 3D estimation methods that make full use of both 2D and 3D ground truths. These fully supervised methods aim to learn the relationship between 2D and 3D data with the help of given paired 2D and 3D data.

Sun et al.\cite{ferrari_integral_2018} propose an end-to-end integral regression model to extract 3D poses from 2D heatmaps. Madadi et al.\cite{MADADI2020107472} use CNN-based 3D joint predictions as an intermediate representation to regress the SMPL pose and shape parameters, and then reconstruct 3D joints in the SMPL output.
Dushyant et al. \cite{2017VNect} propose a method utilizing a fully convolutional network, which regresses 2D and 3D joint positions and motion skeletons to produce a real-time stable 3D reconstruction of motion.
Different from the end-to-end methods, Martinez et al.\cite{2017A} use a simple but effective regression network to learn the correspondences from 2D poses to 3D poses without using any image information.
Moreno-Noguer\cite{20173D} implements an approach to learn the correspondence between the 2D distance matrix and 3D distance matrix with a regression model.
Wang et al.\cite{ijcai2018-136} use 3D data to train an intermediate ranking network and estimate 3D poses from 2D poses by predicting the depth rankings of human joints.
Yuan et al.\cite{yuan2021simpoe} propose a simulation-based approach that integrates image-based kinematic inference and physics-based dynamics modeling to improve the accuracy of video-based 3D human pose estimation.
Zhan et al.\cite{yzhan2022} convert the input from pixel space to 3D normalized rays, which makes their method robust to camera intrinsic parameter changes. They explicitly take the camera extrinsic parameters as input and jointly model the distribution between the 3D pose rays and camera extrinsic parameters to deal with the in-the-wild camera extrinsic parameter variations, which significantly improves the performance of their model.
\subsection{Unsupervised methods}
Unlike supervised methods, unsupervised 3D estimation methods do not involve the 3D ground truth during the training process.

Rhodin et al.\cite{2018Unsupervised1} propose an encoder-decoder to estimate 3D poses based on unsupervised geometry-aware representations. Multiple 2D projections are required to apply a multiview consistency constraint to learn the appearance representation.
Yasunori Kudo et al.\cite{2018Unsupervised} design a GAN whose generator uses the $x$ and $y$ coordinates of important joint points as input and outputs the predicted value of the $z$-direction component. They assume that if the predicted 3D human body is correct, the 2D reprojection should not collapse even if the 3D human body is rotated at any angle around the $y$-axis and then projected onto the $x$-$z$ plane.
Chen et al.\cite{2020Unsupervised} also design an unsupervised GAN to estimate 3D poses; half of their model is based on a similar strategy and reprojects the generator's output 3D estimation back to the 2D reprojection, which is used as the input of the discriminator. The other half of their model lifts the 2D reprojection to a 3D pose again and reprojects this 3D pose back to 2D once more. Wandt et al.\cite{wandt2021canonpose} use multi-view 2D poses to estimate 3D poses and cameras to impose multi-view consistency and camera consistency to improve the accuracy of unsupervised 3D human pose estimation. Schmidtke et al.\cite{schmidtke2021unsupervised} propose a method to estimate human-interpretable landmarks by transforming a template consisting of predefined body parts that are characterized by 2D Gaussian distributions. Deng et al.\cite{deng2021svmac} use a single 2D pose as input to estimate the 3D pose and camera, and then reproject the 3D pose to multiple random angles to impose single-view-multi-angle consistency to improve the performance of their model.

\subsection{Weakly supervised methods}
Weak supervision only requires limited 3D labels or an unpaired 2D-3D correspondence.
Zhou et al.\cite{2017Towards} propose a two-stage transfer model to generate 2D heatmaps and regress the joint depths to estimate 3D poses. The 2D and 3D data are mixed during the training process. Hsiao-Yu Fish Tung et al.\cite{2017Adversarial} propose an adversarial inverse graph network model. This model uses the presentation feedback of the prediction results to map the image to the latent factors and matches the distribution between the predicted results and the ground truth latent factors.

Recently some weakly supervised methods have been proposed based on adversarial architectures and reprojection constraints, that is, an estimated 3D pose should be correctly projected back to the 2D pose.
Yang et al.\cite{yang_3d_2018} implement an adversarial architecture based on multiple representations, including RGB images, geometric representations and heatmaps to estimate 3D poses from in-the-wild 2D images.
Bastian Wandt et al.\cite{2019RepNet} propose a GAN-based model named RepNet to learn a mapping from a distribution of 2D poses to a distribution of 3D poses with an adversarial training approach, in which a camera estimation network is a part of the generator. Li et al.\cite{LI2020BMVC} design a network to model a proposal distribution to approximate the unknown multi-modal target posterior distribution. This approximation is obtained by minimizing the KL divergence between the proposal and target distributions, leading to 2D reprojection error and a prior loss term that makes their model weakly supervised. Xu et al.\cite{Xu2021TPAMI} propose a resolution-aware network RSC-Net, which uses a contrastive learning scheme to learn 3D body pose and shape across different resolutions with one single model. The contrastive learning scheme enforces scale-consistency of the deep features.

In summary, most recent unsupervised and weakly supervised methods involving adversarial structures only consider the consistency constraints between 2D poses and 2D reprojections or among several lifted 3D poses\cite{2020Unsupervised}.
In this paper, we propose a weakly supervised method that considers 3D estimations along with 2D reprojections simultaneously and trains a reprojection network with a GAN synchronously. We also propose a weighted KCS matrix and use it as one part of the discriminator's input to improve the 3D pose estimation accuracy. The experimental results show that our model outperforms state-of-the-art methods.

\section{Methods}
\begin{figure*}[tbp]
\centering
\includegraphics[width=0.9\textwidth]{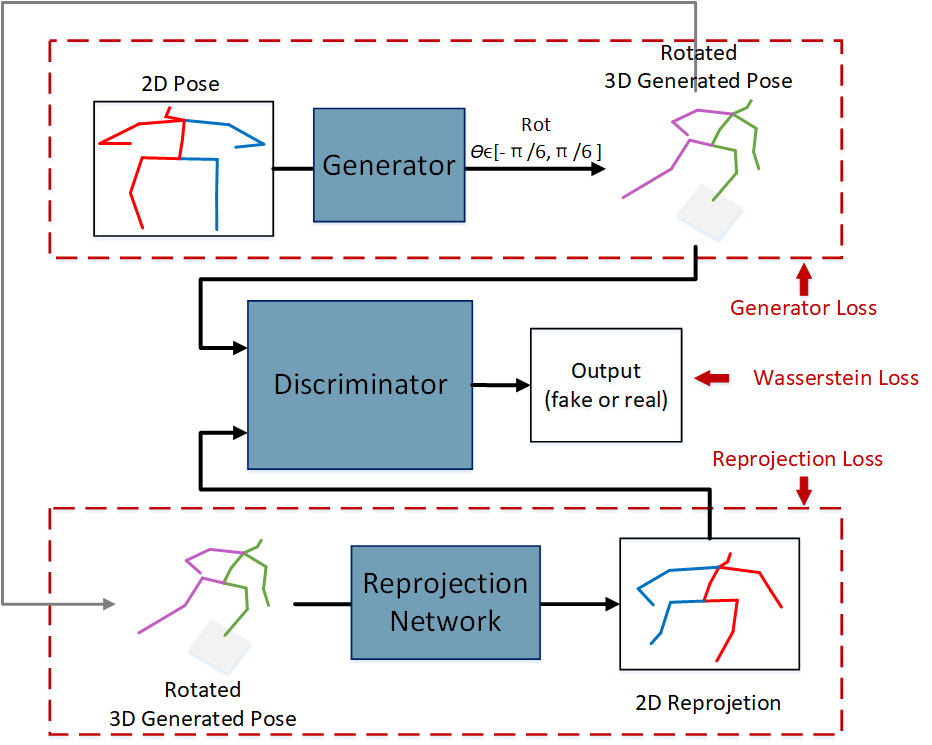}
\caption{The main structure of our proposed adversarial training framework, which contains 4 parts: (1) generator, (2) discriminator, (3) reprojection network, and (4) three loss functions. Our structure takes 2D reprojected poses and 3D generated poses into consideration simultaneously. In pratice, the generator, discriminator and reprojection network will be trained synchronously.}
\label{netStru}
\end{figure*}

\begin{figure*}[tbp]
\centering
\includegraphics[width=1\textwidth]{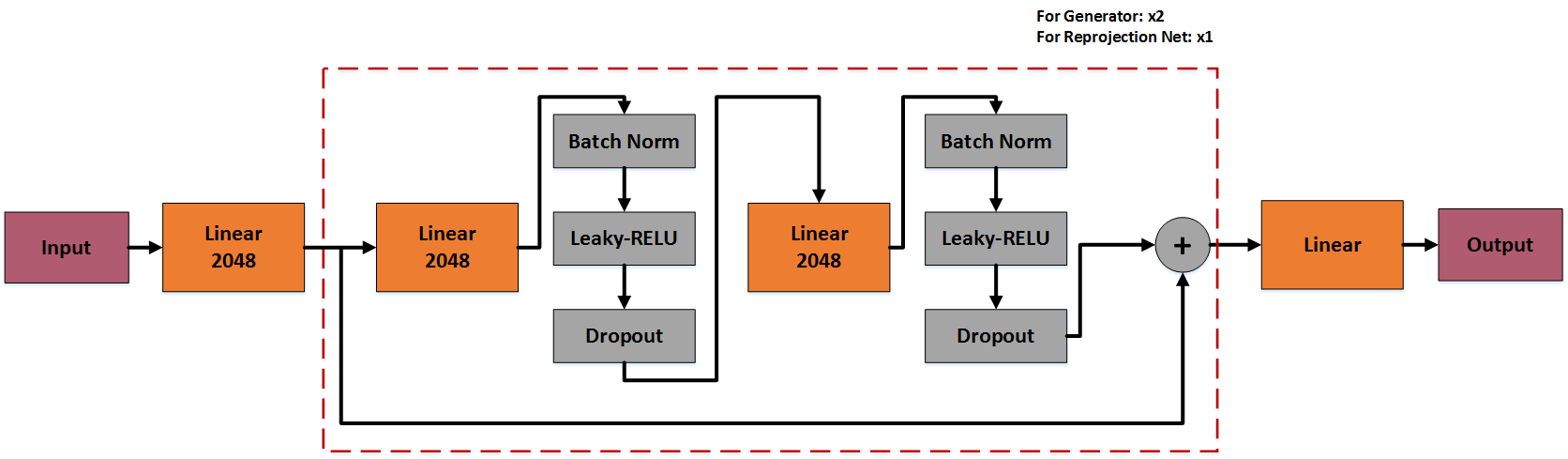}
\caption{The structure of our generator and reprojection network is roughly the same as Martinez\cite{2017A}. The generator has two residual blocks, and the reprojection network has only one residual block. The generator's input is a 2D pose, and the output is the corresponding $z$-direction component. In contrast, the reprojection network's input is a 3D pose, and the output is the related reprojected 2D pose.}
\label{GenRep}
\end{figure*}
For a given 2D pose, our goal is to estimate its corresponding 3D pose. The framework we use can be formulated as a GAN.
In standard GAN training, the generator's input is a Gaussian distribution or a uniform distribution. The discriminator is used to determine whether the input data are from the real distribution or generated by the generator. The generator and discriminator are trained alternately. Finally, alternate training makes the generator's output increasingly closer to the distribution of the real data.

Generally, the input of a standard GAN is sampled from a Gaussian or uniform distribution. In our network architecture, similar to RepNet\cite{2019RepNet}, we assume that the generator's input is sampled from a distribution of 2D observations of human poses (including the $x$ and $y$ coordinates) obtained from the RGB images. The generator generates reasonable $z$-components of the 3D human poses without the camera parameters. However, due to its randomness, it is highly probable that the generated 3D pose is far from the real 3D pose. Hence, we employ more constraints to improve the generator's performance to produce a more realistic 3D pose. Our network architecture is shown in Figure \ref{netStru}.

\subsection{Generator}
The input of our generator is 2D joint coordinates $X_{real}\in \mathbb{R}^{2N}$,
where $N$ represents the number of joints.
The generator estimates the depth of the 2D input, and then we can obtain the corresponding 3D poses $Y_{pred}\in \mathbb{R}^{3N}$.
Our generator is designed to learn the mapping from a 2D distribution to a 3D distribution. Its network architecture is shown in Figure \ref{GenRep} and is similar to Martinez's architecture\cite{2017A}, which has two residual blocks\cite{2016Deep}, each of which contains two hidden layers, batch norms\cite{2015Batch}, leaky-ReLU\cite{b_xu_n_wang_t_chen_m_li_empirical_2015}, dropout\cite{2014Dropout}, etc.

\subsection{Reprojection network}
To produce a more realistic 3D pose, we impose reprojection constraints; that is, the generated 3D pose can still be reprojected back to the original 2D pose. We first obtain the rotated 3D generated pose $Y_{rot}$ by rotating $Y_{pred}$ around the $y$-axis by a small random angle that is obtained from a uniform distribution on $[-\pi, \pi]$, which imposes the constraints that the generated 3D pose should remain reasonable when viewed by any angle. Then, we use a reprojection network to impose reprojection constraints.
Our reprojection network's input is the rotated 3D generated pose $Y_{rot}\in \mathbb{R}^{3N}$, and the output is the reprojected 2D pose $X_{rep}\in \mathbb{R}^{2N}$. Instead of training it alone, we use the generated 3D pose together with the reprojected 2D pose as two parts of the discriminator's input and train the three networks synchronously so that our reprojection net can learn the mapping from the distribution of 3D poses to 2D poses instead of a simple correspondence. In this way, our reprojection net can provide more reliable weakly supervised information to train the generator. Through many experiments, we find that the result of this synchronous training strategy is better than the result obtained by training the reprojection network alone. The structure of our reprojection network is shown in Figure \ref{netStru}. It is similar to the structure of our generator but only includes one residual block.

\begin{figure*}[tbp]
\centering
\includegraphics[width=0.8\textwidth]{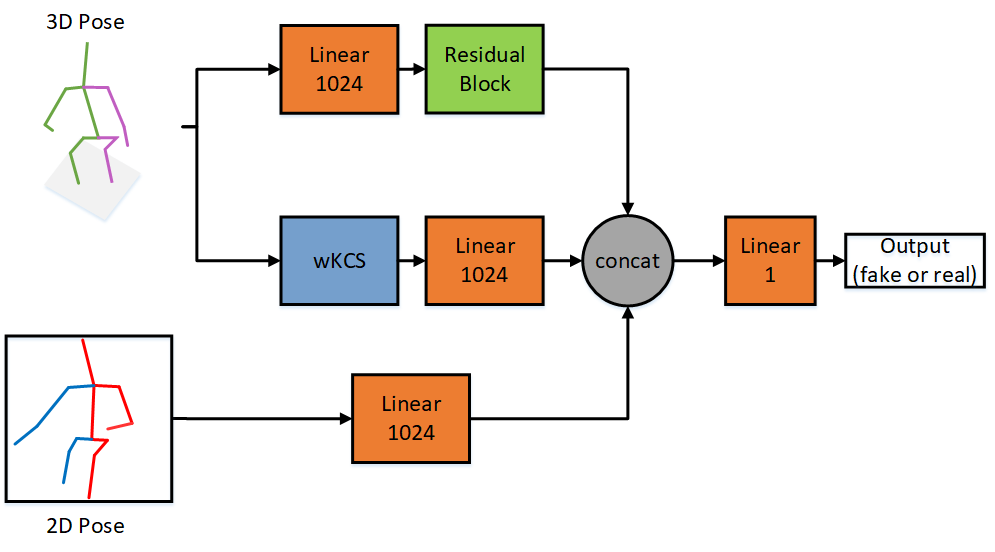}
\caption{The structure of the discriminator.}
\label{Disc}
\end{figure*}

\subsection{Discriminator}
Our discriminator network architecture is shown in Figure \ref{Disc}, and its input has three parts: the 3D pose, 2D pose, and weighted KCS matrix. 
The 3D pose part is the generated or the ground truth 3D poses, and the 2D pose part is the reprojected or ground truth 2D poses. In the following, we describe the weighted KCS matrix.

The original KCS matrix, first proposed by Wandt et al.\cite{2019RepNet}, is used to project 3D poses into kinematic chain space to impose constraints on bone lengths and joint angles.
In the following, we simply describe the calculation method for the original KCS matrix. First, we define a piece of bone $b_k$ as the vector between the $r$-th joint and the $t$-th joint,
$$b_k=p_r-p_t=Yc,\eqno{(1)}$$
where $Y$ represents a 3D pose and 
$$c=(0,\ldots,0,1,0,\ldots,0,-1,0,\ldots,0)^T.\eqno{(2)}$$
$c$ has $j$ terms, and the terms 1 and -1 represent the start and end indexes of a bone vector, respectively. Let $C\in \mathbb{R}^{j\times b}$ contain the start and end indexes of all bones, where $b$ represents a 3D pose with $b$ bones. Hence, we have
$$B=(b_1,b_2,\ldots,b_b)=YC,\eqno{(3)}$$
and we obtain the following KCS matrix:
$$KCS=B^TB.\eqno{(4)}$$

However, through the calculation process of the original KCS matrix, we find that the matrix treats the joint angle information between any two bones equally, no matter how far apart the two bones are. Hence, we adopt a weighted kinematic chain space to individually consider the importance of the angle information of each bone to any other bone. The main idea is that for 3D human pose estimation, the closer the two bones are, the more critical their angle information is.
In practice, we transform the 3D human poses into a weighted KCS and make it one part of the discriminator's input.
\begin{figure*}[tbp]
\centering
\includegraphics[width=0.25\textwidth]{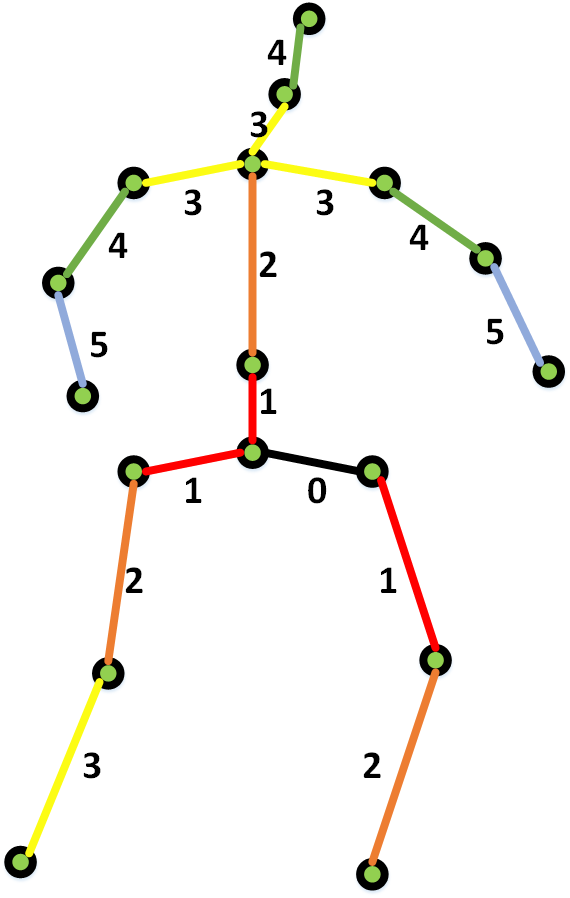}
\caption{The bone distances $d_{ij}$ from the right hip bone.}
\label{iKCS}
\end{figure*}

After obtaining the original KCS matrix, we determine the weight $w_{ij}$ of each element of the KCS. We define a bone distance $d_{ij}$ to represent the distance between the $i$-th bone and the $j$-th bone. For example, in Figure \ref{iKCS}, we give the bone distances from the right hip bone, where the distance to itself is defined as 0.
There are three bones whose distances are 1 and two bones whose distances are 5.
By the bone distances of each bone to any other bone, the weight $w_{ij}$ of the KCS matrix is calculated as follows:
$$w_{ij} = \left\{
\begin{array}{ll}
1,  &    i=j \\
1,  & d_{ij} = 1 \\
tanh(\frac{1}{(d_{ij}-1)}), & d_{ij} > 1
\end{array}
\right.\eqno{(5)}$$
In the definition, for the matrix's diagonal elements, i.e., $i=j$, we retain the bone length information without the weights. For each bone's adjacent bones, i.e., $d_{ij}=1$, their knowledge of the joint angles do not change. Let $W\in \mathbb{R}^{b\times b}$ contain all the items of $w_{ij}$; then, we obtain the following weighted KCS matrix:
$$wKCS = W * KCS,\eqno{(6)}$$
where $*$ represents an element-wise multiplication. In practice, the weighted KCS matrix is easy to calculate. Since the wKCS contains information on bone lengths and joint angles, it is more convenient for optimizing the generator to obtain a more proper 3D human pose.

Back to our discriminator's input, the 3D pose part can create a feature vector through a fully connected layer and a residual block, the 2D pose part generates a feature vector through a fully connected layer, and the weighted KCS part also generates a feature vector through a fully connected layer. The three feature vectors have the same dimension. They are then concatenated and fed into a fully connected layer, which generates the discriminator's output. Let
$$p_{real}=D^*(Y_{real},X_{real},wKCS_{real}),\eqno{(7)}$$
$$p_{fake}=D^*(Y_{rot},X_{rep},wKCS_{pred}),\eqno{(8)}$$
where $wKCS_{pred}$ represents the weighted KCS matrix of the rotated 3D generated pose $Y_{rot}$, and $wKCS_{real}$ represents the weighted KCS matrix of the ground-truth 3D pose $Y_{real}$. $p_{real}$ is the discriminator's output when the ground truth 3D poses and the ground truth 2D poses are fed to the discriminator and $p_{fake}$ is the discriminator's output when the generated 3D poses and the reprojected 2D poses are the input.

\begin{figure*}[tbp]
\centering
\includegraphics[width=1.0\textwidth]{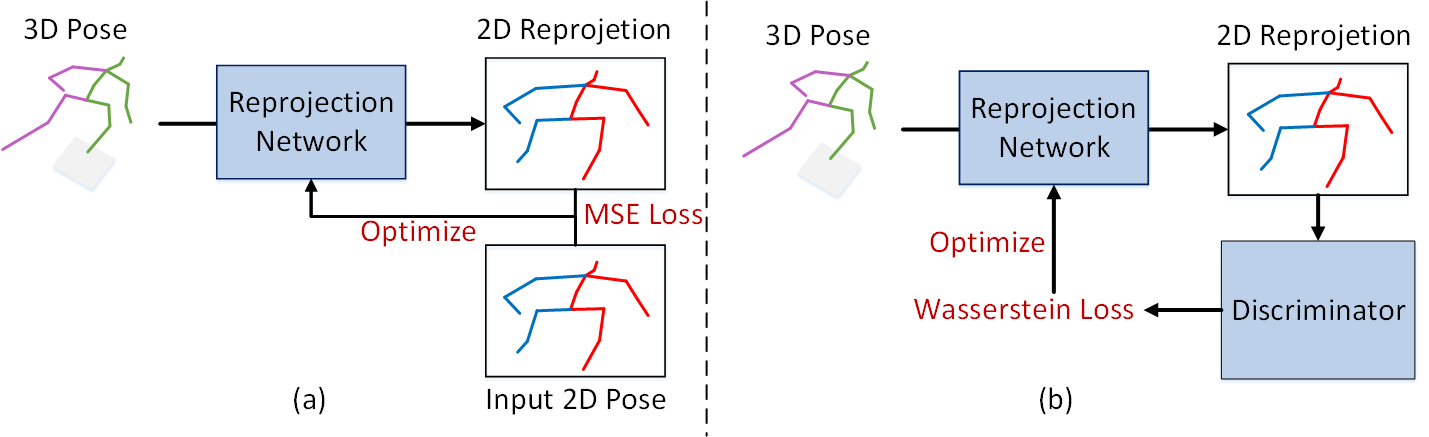}
\caption{The difference between ordinary reprojection methods (left) and our synchronous training strategy (right). (a) Many methods calculate the MSE loss between the reprojected 2D poses and the input 2D poses to optimize the reprojection network; (b) We input the 2D reprojection to the discriminator and use the wasserstein loss of GAN to optimize the reprojection network.}
\label{diff}
\end{figure*}

\subsection{Synchronized training strategy}
Many methods\cite{2019Lifting, 2019RepNet} impose reprojection constraints by comparing the reprojected 2D pose with the original input 2D pose and calculating the error to optimize the reprojection network. However, such reprojection methods are usually influenced by the input 2D poses; that is, when they are detected by a 2D human pose estimator instead of the ground truth, the 2D reprojection accuracy will be correspondingly reduced.

To overcome such shortages, we design a synchronous training strategy to optimize the three neural networks in our adversarial training model. In practice, we synchronously train the generator, the reprojection network, and the discriminator using the loss function of the generative adversarial network.
Figure \ref{diff} shows the difference between our synchronous training strategy and conventional reprojection methods.
With the help of our synchronous training strategy, the generator learns the mapping from the 2D distribution to the 3D distribution, while the reprojection network learns the mapping from the 3D distribution to the 2D distribution. As a result, the reprojection network can provide more reasonable weakly supervised information for the generator and reduce the influence of errors in the input of the generator, which can help train the generator better.

\subsection{Loss functions}
First, for the GAN, we use the Wasserstein loss function\cite{2017WGAN}, i.e.,
$$L_{dis}=p_{fake}-p_{real}.\eqno{(9)}$$
To train our three networks synchronously, the loss function of the generator is the same as that of the reprojection network,
$$L_{gen}=L_{rep}=-p_{fake}.\eqno{(10)}$$
Then, we impose another constraint $L_{angle}$, which guarantees that the $z$-components of the generated 3D pose will not be inverted, by referring to Yasunori\cite{2018Unsupervised}. Similarly, we define the face orientation vector
$v=[v_x,v_y,v_z ]=j_{nose}-j_{neck}\in R^3$ and shoulder orientation vector $w=[w_x,w_y,w_z]=j_{ls}-j_{rs}\in \mathbb{R}^3$, where $j_{nose},j_{neck},j_{ls},j_{rs}\in \mathbb{R}^3$ represent the 3D coordinates of the nose, neck, left shoulder and right shoulder respectively. According to the abovementioned constraints, the angle $\beta$ between $v$ and $w$ on the $z-x$ plane should satisfy
$$\sin\beta =\frac{v_z w_x-v_x w_z}{\| v\|\| w\|}\geq 0.\eqno{(11)}$$
To satisfy this inequality, let
$$L_{angle}=max(0,-\sin\beta)=max(0,\frac{v_x w_z-v_z w_x}{\| v\|\| w\|} ).\eqno{(12)}$$

Finally, through equations (10) and (12), we obtain the final loss function of the generator as follows:$$L_{gen}=-p_{fake}+\lambda L_{angle},\eqno{(13)}$$
where $\lambda$ represent the weight coefficients of the loss terms $L_{angle}$.

\subsection{Data processing}
We performed data preprocessing on the data before training. Similar to most pose estimation methods, we use the human hip joint as the root joint and subtract the coordinates of the other joint points from the root joint by translating them relative to the root node. Then, we divide the value of all joint coordinates by the corresponding ratio, which is the average of the Euclidean distances from all joints to the root joint. In the training and testing phases, we use these coordinates to represent each joint's position.

\subsection{Training details}
As mentioned above, we use the standard Wasserstein GAN (WGAN) loss function and our loss function $L_{angle}$ to train our generator, reprojection network, and discriminator synchronously during each iteration. In the following experiments, the input ($Y_{real}$, $X_{real}$) corresponds to the 3D-2D label of ($Y_{pred}$, $X_{rep}$). Due to the utilization of GAN during the training process, we do not directly compute the loss based on ground-truths and outputs. We use the Adam optimizer for all three networks with a learning rate of 8e-5, $beta_1$ = 0.0 and $beta_2$ = 0.9. The loss weights are set as $\lambda$ = 1. 

\section{Experiments}
{\bf Dataset} We perform experiments on the three datasets, Human3.6M\cite{2014Human3}, MPI-INF-3DHP\cite{2017Monocular} and MPII\cite{2014Human}. Human3.6M is the most popular benchmark dataset for 3D human pose estimation and contains over 3.6 million 3D human poses and the corresponding images. To compare our results with the results of state-of-the-art methods in related fields, we also use the 2D poses estimated by the stacked hourglass\cite{2016Stacked} method on the Human3.6M\cite{2014Human3} dataset to perform 3D estimation experiments. Finally, we test our model on the MPII dataset, and the experimental results show that our model performs well, even on the dataset whose images are captured from a monocular camera.

\subsection{Quantitative evaluation on Human3.6M}
{\bf Protocols} For the Human3.6M dataset, we use S1, S5, S6, S7, and S8 as the training sets and S9 and S11 as the test sets. The evaluation standard is the mean per joint positioning error (MPJPE). The MPJPE calculates the average Euclidean distance between the estimated 3D pose and the ground truth 3D pose. In the experiments, there are two main protocols to follow: protocol \#1 calculates the MPJPE directly and protocol \#2 calculates the MPJPE after aligning the estimation with the ground truth via a rigid transformation\cite{Bogo2016Keep, 20173D}.

\begin{table*}[tbp]
\tiny
\setlength{\abovecaptionskip}{0pt}
\setlength\tabcolsep{0pt}
\centering
\caption{The results of 3D human pose estimation of the Human3.6M dataset compared to other state-of-the-art methods following Protocol \#1, all referred results come from the related papers. 'GT2D' indicates whether the method uses ground-truth 2D poses as input.'WS' indicates whether the method is weakly supervised.}\smallskip
\begin{tabular}{lclllllllllllllllll}
\hline
Protocol \#1 &GT2D &WS & Direct & Discuss & Eating &Greet &Phone &Photo &Pose &Purch &Sitting &SittingD &Smoke &Wait &WalkD &Walk &WalkT &Avg\\
\hline
Martinez et al.\cite{2017A}&\checkmark
	&& 37.7& 44.4& 40.3& 42.1& 48.2& 54.9& 44.4& 42.1& 54.6& 58.0& 45.1& 46.4& 47.6& 36.4& 40.4& 45.5\\
Xu et al.\cite{2020Deep}&\checkmark&&40.6&47.1&45.7&46.6&50.7&63.1&45.0&47.7&56.3&63.9&49.4&46.5&51.9&38.1&42.3&49.2\\
Xu et al.\cite{xu2021graph}&\checkmark&& 35.8& 38.1& 31.0& 35.3& 35.8& 43.2& 37.3& 31.7& 38.4& 45.5& 35.4& 36.7& 36.8& 27.9& 30.7& 35.8\\
RepNet\cite{2019RepNet}&
\checkmark	&\checkmark&50.0	&53.5	&44.7	&51.6	&49.0	&\textbf{58.7}	&48.8	&51.3	&51.1	&66.0	&46.6	 &50.6	 &\textbf{38.8} &\textbf{42.5}	&60.4	&50.9\\
\textbf{Ours (GT)}&\checkmark&\checkmark&\textbf{43.3}&\textbf{52.9}&\textbf{42.5}&\textbf{48.7}&\textbf{45.9}&59.8&\textbf{44.9}&\textbf{45.9}&\textbf{49.5}&\textbf{55.9}&\textbf{44.9}&\textbf{49.6}&51.3&43.0&\textbf{45.9}&\textbf{48.3}\\
\hline
LinKDE\cite{2014Human3}&&
	&132.7	&183.6	&132.3	&164.4	&162.1	&205.9	&150.6	&171.3	&151.6	&243.0	&162.1	&170.7	&177.1	 &96.6	 &127.9	&162.1\\
Du et al.\cite{2016Marker}&&
	&85.1	&112.7	&104.9	&122.1	&139.1	&135.9	&105.9	&166.2	&117.5	&226.9	&120.0	&117.7	&137.4	 &99.3	 &106.5	&126.5\\
Zhou et al.\cite{zhou_sparseness_2016}&&
	&87.4	&109.3	&87.1	&103.2	&116.2	&143.3	&106.9	&99.8	&124.5	&199.2	&107.4	&118.1	 &79.4 &114.2	 &97.7	&113.0\\
Martinez et al.\cite{2017A}&
	&&51.8& 56.2& 58.1& 59.0& 69.5& 78.4& 55.2& 58.1& 74.0& 94.6& 62.3& 59.1& 65.1& 49.5& 52.4& 62.9\\
Kudo et al.\cite{2018Unsupervised}&&
	&125.0	&137.9	&107.2	&130.8	&115.1	&127.3	&147.7	&128.7	&134.7	&139.8	&114.5	&147.1	&130.8	 &125.6	 &151.1	&130.9\\
Coarse2Fine\cite{2017Coarse}&&
	&67.4	&71.9	&66.7	&69.1	&72.0	&77.0	&65.0	&68.3	&83.7	&96.5	&71.7	&65.8	&74.9	 &59.1	 &63.2	&71.9\\
Xu et al.\cite{xu2021graph}&&&45.2 &49.9& 47.5& 50.9& 54.9& 66.1& 48.5& 46.3& 59.7& 71.5& 51.4& 48.6& 53.9& 39.9& 44.1& 51.9\\
RepNet\cite{2019RepNet}&&\checkmark&77.5&	85.2&	82.7&	93.8&	93.9&	101.1&	82.9&	102.6&100.5&	125.8&	88.0&	84.8&	78.8& 72.6&	79.0&	89.9\\
Li et al\cite{LI2020BMVC}&&\checkmark&67.9&	75.5&	71.8&	81.8&	81.4&	93.7&75.2&81.3&88.8&114.1&75.9&79.1&83.3&74.3&79.0&81.1\\
Xu et al.\cite{Xu2021TPAMI}&&\checkmark&-&-&-&-&-&-&-&-&-&-&-&-&-&-&-&79.6\\
\textbf{Ours (SH)}&&\checkmark&\textbf{62.8}&	\textbf{64.8}&	\textbf{61.2}&	\textbf{66.8}&	\textbf{67.6}&	\textbf{83.5}&	\textbf{62.5}&	\textbf{66.0}&\textbf{70.9}&	\textbf{90.1}&	\textbf{64.3}&	\textbf{65.6}&	\textbf{73.0}&	\textbf{55.1}&	\textbf{61.3}&	\textbf{67.7}\\
\hline
\end{tabular}
\label{P1}
\end{table*}
\begin{figure}[htbp]
\centering
\includegraphics[width=1\textwidth]{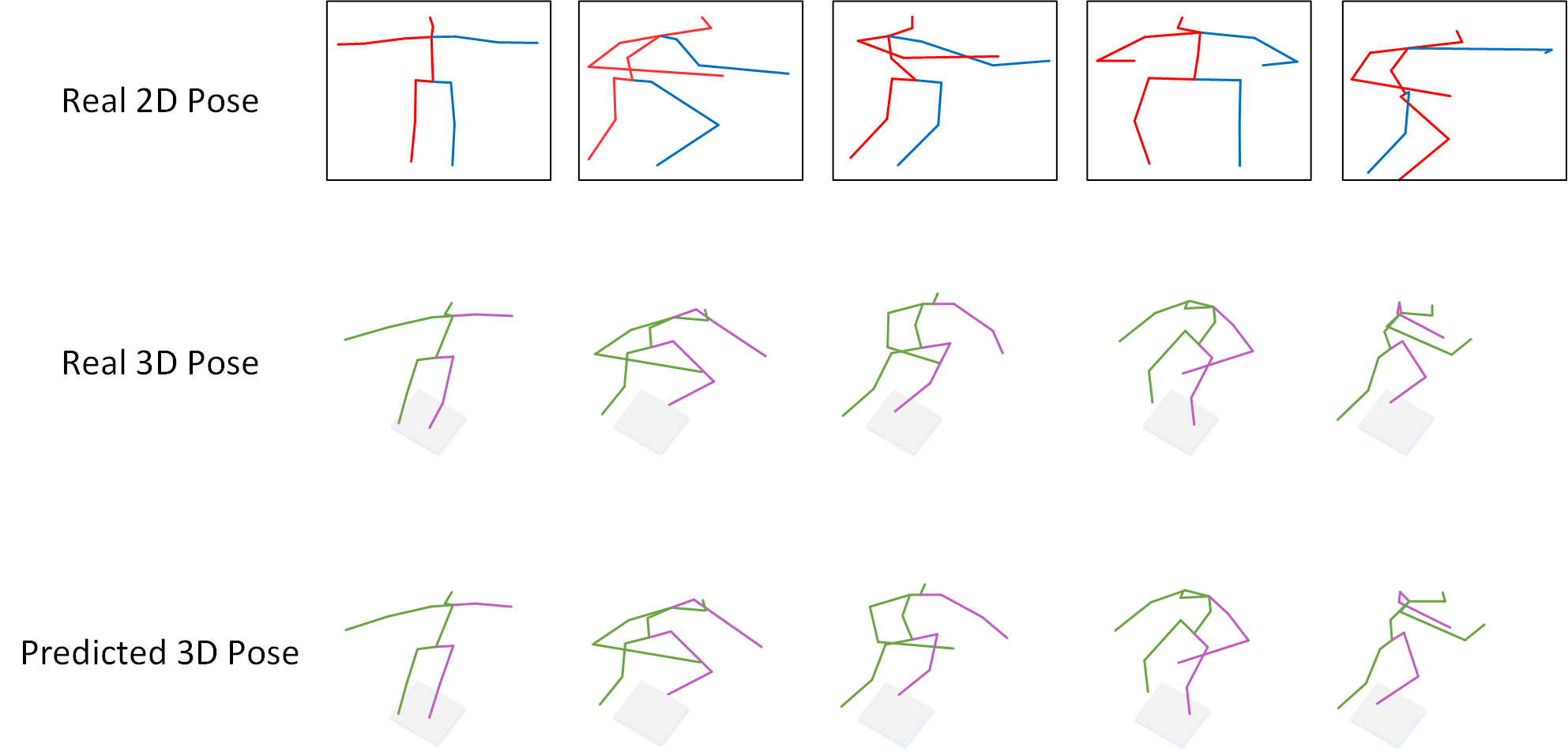}
\caption{Some reconstruction results on the Human3.6M dataset. The first row is ground truth 2D poses, the second row is ground truth 3D poses, and the third row is the reconstructed 3D poses predicted by our model.}
\label{results}
\end{figure}

{\bf Results under protocol \#1} The experimental results following protocol \#1 are shown in Table 1 and include two parts. In the first part, we compare our model with other methods when using ground-truth 2D poses as input. 
It can be seen that our model outperforms the state-of-the-art method RepNet in most actions and achieves an average MPJPE of 48.3 mm, which exceeds RepNet by approximately 5.1\% and is even better than the fully supervised method proposed by Xu et al.\cite{2020Deep}. In the second part,
we conduct the comparison when using detected 2D poses as input. 
It can be seen that our model outperforms the state-of-the-art method RepNet in all actions and achieves an average MPJPE 67.7 mm,
which shows that our model outperforms SOTA methods by approximately 14.9\%. This significant improvement shows that the synchronous training strategy can help the reprojection network generate accurate 2D reprojections, thereby further improving the 3D pose estimation performance of our model.
In addition, our model is even better than several fully supervised methods.
Figure \ref{results} shows some reconstruction examples on the Human 3.6M dataset, and it can be seen that even difficult poses can be reconstructed well. 

{\bf Results under protocol \#2} The experimental results under protocol \#2, which uses a rigid alignment with the ground-truth 3D poses, are shown in Table 2 and include two parts. In the first part, we compare our model with other methods when using ground-truth 2D poses as input. The results show that our model obtains stable and balanced pose estimations, and outperforms the state-of-the-art weakly supervised methods by approximately 9.9\%. In the second part,
we conduct the comparison when using detected 2D poses as input. In this part, our method outperforms most SOTA methods. Compared with \cite{LI2020BMVC}, our method obtains the second-best results in terms of the average P-MPJPE and the best results in two actions. Additionally, our method is better than several fully supervised methods.

{\bf Robustness to detector noise} Similar to \cite{20173D}, we add Gaussian noise to the ground truth 2D poses to train and test our model. The mean value of the Gaussian noise is 0, and its standard deviation values are 5, 10, 15, and 20. We perform three groups of experiments, 
as shown in Table 3. The top part includes the results obtained by applying Gaussian noise to training and testing and then calculating the MPJPE. The middle part contains the results obtained by applying Gaussian noise only to the training data but testing on ground truth and then calculating the MPJPE. 
The bottom part includes the
results obtained by applying Gaussian noise to the training and test data and then calculating the $z$-component error only.
We perform all three experiments to evaluate our model's ability to estimate the depth of human poses and to verify its reliability. The experimental results show that our model can still perform well in estimating the human pose depth even if the 2D detector produces noise.

\begin{table*}[tbp]
\tiny
\setlength{\abovecaptionskip}{0pt}
\setlength\tabcolsep{0pt}
\centering
\caption{The results of 3D human pose estimation on the Human3.6M dataset compared with other state-of-the-art methods following protocol \#2;  all referenced results come from the related papers. 'GT2D' indicates whether the method uses ground-truth 2D poses as input. 'WS' indicates whether the method is weakly supervised.}\smallskip
\begin{tabular}{lclllllllllllllllll}
\hline
Protocol \#2 & GT2D &WS& Direct & Discuss & Eating &Greet &Phone &Photo &Pose &Purch &Sitting &SittingD &Smoke &Wait &WalkD &Walk &WalkT &Avg\\
\hline
Xu et al.\cite{2020Deep}&\checkmark&&33.6&37.4&37.0&37.6&39.2&46.4&34.3&35.4&45.1&52.1&40.1&35.5&42.1&29.8&35.3&38.9\\
3Dinterpreter\cite{wu2016single}&\checkmark&\checkmark&56.3& 77.5& 96.2& 71.6& 96.3& 106.7& 59.1& 109.2& 111.9& 111.9& 124.2&93.3&-& 58.0&-& 88.6\\
AIGN\cite{2017Adversarial}&\checkmark&\checkmark&53.7& 71.5& 82.3& 58.6& 86.9& 98.4& 57.6& 104.2& 100.0& 112.5& 83.3& 68.9&-& 57.0& -&79.0\\
RepNet\cite{2019RepNet}&
\checkmark&\checkmark&33.6&38.8&32.6&37.5&36.0&44.1&37.8&34.9&39.2&52.0&37.5&39.8&40.3&34.1&34.9&38.2\\
\textbf{Ours (GT)}&\checkmark&\checkmark&\textbf{32.1}&\textbf{36.6}&\textbf{30.1}&\textbf{36.3}&\textbf{31.4}&\textbf{39.6}&\textbf{33.4}&\textbf{31.4}&\textbf{34.5}&\textbf{40.8}&\textbf{33.5}&\textbf{34.8}&\textbf{36.1}&\textbf{30.9}&\textbf{34.7}&\textbf{34.4}\\
\hline
Zhou et al.\cite{2016Sparse}&&
&99.7&95.8&87.9&116.8&108.3&107.3&93.5&95.3&109.1&137.5&106.0&102.2&110.4&106.5&115.2&106.7\\
Bogo et al.\cite{Bogo2016Keep}&
&&62.0&60.2&67.8&76.5&92.1&77.0&73.0&75.3&100.3&137.3&83.4&77.3&79.7&86.8&87.7&82.3\\
Martinez et al.\cite{2017A}&
&&39.5&43.2&46.4&47.0&51.0&56.0&41.4&40.6&56.5&69.4&49.2&45.0&49.5&38.0&43.1&47.7\\
Fang et al.\cite{2017Learning}&
&&38.2&41.7&43.7&44.9&48.5&55.3&40.2&38.2&54.5&64.4&47.2&44.3&47.3&36.7&41.7&45.7\\
3Dinterpreter\cite{wu2016single}&&\checkmark&78.6& 90.8& 92.5& 89.4& 108.9& 112.4& 77.1& 106.7& 127.4& 139.0& 103.4& 91.4&- &79.1 &- &98.4\\
AIGN\cite{2017Adversarial}&&\checkmark&77.6& 91.4& 89.9& 88.0& 107.3& 110.1& 75.9& 107.5& 124.2& 137.8& 102.2& 90.3& -&78.6& -& 97.2\\
RepNet\cite{2019RepNet}&&\checkmark&53.0&	58.3&	59.6&	66.5&	72.8&	71.0&	56.7	&69.6& 78.3&	95.2&	66.6&	58.5&	57.5& 63.2&	49.9&	65.1\\
Li et al.\cite{LI2020BMVC}&&\checkmark&\textbf{42.1}&	\textbf{44.7}&	\textbf{45.4}&	\textbf{51.0}&	\textbf{49.3}&	\textbf{51.5}&	\textbf{41.2}	&\textbf{46.2}& 57.5&	\textbf{70.8}&	\textbf{48.7}&	\textbf{44.1}&	\textbf{50.8}& 42.1&	\textbf{43.7}&	\textbf{48.7}\\
\textbf{Ours (SH)}&&\checkmark&48.8&48.4&46.6&52.3&50.9&61.2&48.1&46.3&\textbf{55.5}&71.1&51.0&48.5&54.3&\textbf{41.9}&48.7&51.6\\
\hline
\end{tabular}
\label{tab:bike}
\end{table*}

\begin{table*}[tbp]
\tiny
\setlength{\abovecaptionskip}{0pt}
\setlength\tabcolsep{0pt}
\centering
\caption{The robustness of our model to 2D detector noise on the Human3.6M dataset following protocol \#2. 'GT2D' indicates whether the method uses ground-truth 2D poses for testing. 
}\smallskip
\begin{tabular}{lccllllllllllllllll}
\hline
Protocol \#2 & GT2D &Error& Direct & Discuss & Eating &Greet &Phone &Photo &Pose &Purch &Sitting &SittingD &Smoke &Wait &WalkD &Walk &WalkT &Avg\\
\hline
GT
&&MPJPE&32.1&36.6&30.1&36.3&31.4&39.6&33.4&31.4&34.5&40.8&33.5&34.8&36.1&30.9&34.7&34.4\\

GT+N(0, 5)                     &&MPJPE&50.7&51.6&48.9&55.1&51.3&57.5&51.8&49.3&51.9&55.7&51.8&51.1&54.4&47.7&51.9&52.1\\
GT+N(0, 10)                   &&MPJPE&72.3&73.9&71.6&77.8&74.3&80.0&74.6&73.5&75.4&83.7&75.3&75.7&78.0&73.7&76.1&75.7\\
GT+N(0, 15)                   &&MPJPE&96.3&97.1&95.8&100.7&97.1&102.5&96.5&97.1&98.5&106.9&97.5&97.6&101.4&95.8&100.4&98.8\\
GT+N(0, 20)                   &&MPJPE&115.7&116.3&119.9&120.9&118.0&125.3&116.5&116.9&120.7&132.0&117.2&120.4&120.6&119.9&121.2&120.1\\
\hline
GT+N(0, 5)            &\checkmark       &MPJPE&37.5&38.2&35.6&42.9&38.3&46.4&38.9&34.9&39.5&41.9&38.9&38.1&41.6&33.7&38.5&39.0\\
GT+N(0, 10)          &\checkmark       &MPJPE&42.2 &43.1&40.7&48.5&45.4&52.9&45.9&41.2&45.6&53.7&45.2&44.9&47.5&42.9&46.7&45.8\\
GT+N(0, 15)          &\checkmark       &MPJPE&45.9&48.0&46.3&52.1&50.8&58.2&48.8&46.9&51.5&58.5&49.4&48.1&54.6&44.4&52.7&50.4\\
GT+N(0, 20)          &\checkmark       &MPJPE&51.8&50.7&55.1&58.3&53.9&63.1&52.6&49.8&54.5&70.5&51.9&56.9&56.9&57.2&58.9&56.1\\
\hline
GT+N(0, 5) &&                     only z&30.9&32.8&30.4&35.4&33.9&37.6&34.3&30.6&33.8&36.7&34.1&33.1&35.3&29.1&32.1&33.3\\
GT+N(0, 10)&&                    only z&36.3&38.7&36.1&42.2&40.1&44.8&40.6&38.7&40.5&47.9&40.8&41.1&42.7&38.2&40.0&40.6\\
GT+N(0, 15)  &&                  only z&42.3&43.9&41.7&46.8&45.1&48.8&45.1&44.6&46.0&53.6&45.1&45.6&48.3&42.9&46.6&45.8\\
GT+N(0, 20)  &&                  only z&43.9&45.3&48.8&49.8&49.0&54.5&44.8&44.9&51.9&63.4&47.4&50.8&49.5&40.0&48.1&49.4\\
\hline
\end{tabular}
\label{tab:bike}
\end{table*}

\begin{table}[t]\footnotesize
\setlength\tabcolsep{10pt}
  \centering
  \caption{The results of 3D human pose estimation on the MPI-INF-3DHP dataset.}
    \begin{tabular*}{1.0\hsize}{@{}@{\extracolsep{\fill}}lcc@{}}
    \hline
    Methods & 3D PCK & MPJPE \\
    \hline
    Mehta et al.\cite{2017Monocular}&76.5&117.6\\
    VNect\cite{2017VNect}&76.6&124.7\\
    Zhou et al.\cite{2017Towards}&69.2&137.1\\
    OriNet\cite{2018OriNet}&81.8&\textbf{89.4}\\
    Yang et al.\cite{yang_3d_2018}&69.0&-\\
    RepNet\cite{2019RepNet}&82.5&97.8\\
    Iqbal\cite{Iqbal}&79.5&109.3\\
    Li et al.\cite{LI2020BMVC}&79.3&-\\
    Xu et al.\cite{Xu2021TPAMI}&-&115.8\\
    \hline
    Ours&\textbf{86.0}&94.9\\
    \hline
    \end{tabular*}%
  \label{tab:datasets}%
\end{table}%

\subsection{Quantitative evaluation on MPI-INF-3DHP}
We also perform experiments on the MPI-INF-3DHP dataset\cite{2017Monocular}, and the experimental results are shown in Table 4. The higher the 3DPCK value or the lower the MPJPE value is, the better the model performs. 
In terms of the 3DPCK, our method obtains the best results and outperforms the SOTA method \cite{LI2020BMVC} by 8.4\%. In terms of the MPJPE, our method obtains the second-best results.

In summary, our model can be applied to multiple datasets and achieve better performance in human pose estimation.

\subsection{Ablation studies}

In this section, we conduct ablation studies to evaluate the effectiveness of our synchronous training strategy and the weighted KCS. The experimental results on Human3.6M and MPI-INF-3DHP are shown in Table 5, where wKCS represents the weighted KCS, and Sep and Syn represent the separate and synchronous training strategies, respectively.

\begin{table}[ht]\footnotesize
  \centering
  \caption{Ablation studies.}
    \begin{tabular*}{1.0\hsize}{@{}@{\extracolsep{\fill}}lcc@{}}
    \hline
      & Human3.6M &MPI-INF-3DHP \\
      \hline
     Methods & MPJPE &3DPCK / MPJPE\\
    \hline
    Sep + KCS&51.6& 84.4 / 98.9\\
    Sep + wKCS&50.7&84.8 / 98.1\\
    Syn + KCS&49.6&84.5 / 98.2\\
    Syn + wKCS&\textbf{48.3}&\textbf{86.0} / \textbf{94.9}\\
    \hline
    \end{tabular*}%
  \label{tab:datasets2}%
\end{table}%

First, we compare the experimental results of the Syn+wKCS and Sep+wKCS combinations to confirm the efficacy of our synchronous training strategy. The results clearly demonstrate that employing the synchronous training strategy leads to an improvement in MPJPE, with a 4.7\% enhancement on the Human3.6M dataset and a 3.3\% improvement on the MPI-INF-3DHP dataset. These findings unequivocally establish the effectiveness of our synchronous training strategy, affirming that our model achieves more precise pose estimations.

Then, we compare the experimental results of the Syn+wKCS and Syn+KCS combinations to validate the effectiveness of our weighted KCS. When training our model with wKCS instead of the standard KCS, we observed an improvement of 2.6\% in MPJPE on the Human3.6M dataset and 3.4\% on the MPI-INF-3DHP dataset. This demonstrates that our weighted KCS enables the generator to focus on crucial joint angle information, specifically the angles between bones with smaller distances.

\subsection{Qualitative evaluation on MPII dataset}
\begin{figure*}[htbp]
\centering
\includegraphics[width=1.0\textwidth]{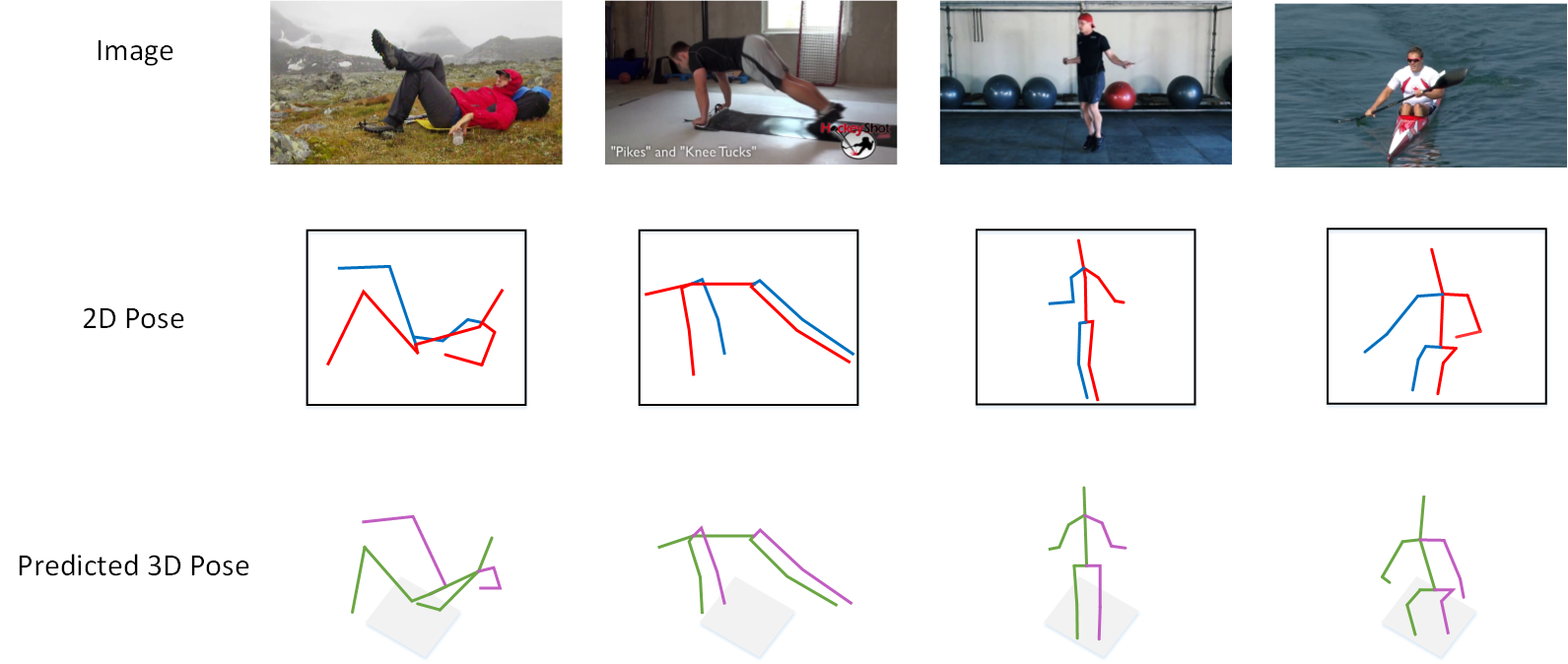}
\caption{Some reconstruction results on the MPII dataset. Each of 2D Pose is the ground truth 2D coordinates, and each of Predicted 3D Pose is the 3D pose predicted using our model. 
}
\label{mpiire}
\end{figure*}

Finally, we conduct experiments on the MPII dataset\cite{2014Human}, which has only 2D annotations. The experimental results are shown in Figure \ref{mpiire}. It can be seen that our model performs well with a standard 2D pose dataset, which contains more complicated in-the-wild poses.

\section{Conclusion}
Many effective models have been developed to estimate 3D human poses. However, most of them only focus on the consistency between 2D poses and 2D reprojections or among several lifted 3D poses to perform this task. In this paper, we propose a synchronized adversarial architecture that utilizes 2D and 3D information simultaneously to estimate 3D human poses from monocular images. Based on a GAN, we add a reprojection network to learn the mapping of the distribution from 3D human poses to 2D reprojections and synchronously train the reprojection network with the generator as well as the discriminator. We also design an improved space that transforms a 3D pose into a weighted kinematic chain space to impose constraints on joint angles and bone lengths.
The experimental results show that our method outperforms the state-of-the-art methods by approximately 24.7\% on Human3.6M and achieves more accurate estimation performance than those methods on MPI-INF-3DHP and MPII. In the future, we plan to improve the model's performance by applying it to multiview images or videos.

\section*{Acknowledgment}
This research is supported by the National Key R\&D Program of China (2021ZD0113002), National Natural Science Foundation of China (No. 62072292, 61572005, 61771058).

%
%

\bibliographystyle{plain}      
\bibliography{reference}   


\end{document}